\documentclass[conference]{IEEEtran}
\IEEEoverridecommandlockouts
\usepackage{cite}
\usepackage{amssymb}
\usepackage{amsmath} 
\usepackage{algorithm} 
\usepackage{algpseudocode} 
\usepackage{amsfonts} 
\usepackage{float}
\usepackage{caption}
\usepackage{xcolor} 
\usepackage{graphicx}
\usepackage{textcomp}
\usepackage{xcolor}
\usepackage{makecell}
\usepackage{multirow}  
\usepackage{array}     
\usepackage{booktabs}  
\usepackage{stfloats}

\usepackage{afterpage}
\usepackage{placeins} 

\def\BibTeX{{\rm B\kern-.05em{\sc i\kern-.025em b}\kern-.08em
    T\kern-.1667em\lower.7ex\hbox{E}\kern-.125emX}}
\begin{document}

\title{VFL-RPS: Relevant Participant Selection in Vertical Federated Learning\\
}


\author{\IEEEauthorblockN{1\textsuperscript{st} Afsana Khan}
\IEEEauthorblockA{\textit{Department of Advanced Computing Sciences} \\
\textit{Maastricht University}\\
Maastricht, Netherlands \\
a.khan@maastrichtuniversity.nl}
\and
\IEEEauthorblockN{2\textsuperscript{nd} Marijn ten Thij}
\IEEEauthorblockA{\textit{Department of Cognitive Science and Artificial Intelligence
} \\
\textit{TilburgUniversity}\\
Tilburg, Netherlands \\
m.c.tenthij@tilburguniversity.edu}
\and
\IEEEauthorblockN{3\textsuperscript{rd} Guangzhi Tang}
\IEEEauthorblockA{\textit{Department of Advanced Computing Sciences} \\
\textit{Maastricht University}\\
Maastricht, Netherlands \\
guangzhi.tang@maastrichtuniversity.nl}
\and
\IEEEauthorblockN{3\textsuperscript{rd} Anna Wilbik}
\IEEEauthorblockA{\textit{Department of Advanced Computing Sciences} \\
\textit{Maastricht University}\\
Maastricht, Netherlands \\
a.wilbik@maastrichtuniversity.nl}
}
\maketitle

\begin{abstract}
Federated Learning (FL) allows collaboration between different parties, while ensuring that the data across these parties is not shared. However, not every collaboration is helpful in terms of the resulting model performance. Therefore, it is an important challenge to select the correct participants in a collaboration. As it currently stands, most of the efforts in participant selection in the literature have focused on Horizontal Federated Learning (HFL), which assumes that all features are the same across all participants, disregarding the possibility of different features across participants which is captured in Vertical Federated Learning (VFL). To close this gap in the literature, we propose a novel method VFL-RPS for participant selection in VFL, as a pre-training step. We have tested our method on several data sets performing both regression and classification tasks, showing that our method leads to comparable results as using all data by only selecting a few participants. In addition, we show that our method outperforms existing methods for participant selection in VFL.
\end{abstract}

\section{Introduction}
Federated learning (FL) is a distributed machine learning paradigm that enables multiple organizations to collaboratively train models without sharing sensitive data, addressing privacy concerns and data protection regulations \cite{kairouz2021advances}. Instead of sharing raw data, FL helps to collaboratively train a global model by exchanging computed updates, such as gradients or parameters, while keeping their data stored locally. FL is categorized into Horizontal Federated Learning (HFL), where datasets share features but differ in samples (e.g., hospitals with similar data types), Vertical Federated Learning (VFL), where datasets share samples but differ in features (e.g., banks and e-commerce platforms with complementary data), and Hybrid Federated Learning, which combines both partitions ~\cite{khan2022vertical}. 

However, FL presents several challenges, including communication overhead, optimal participant selection, ensuring privacy and security during data exchange, addressing fairness among participants, and maintaining robust model performance in the presence of data and participant heterogeneity. In this paper, we focus on the participant selection problem, as including all participants in a federated setting without any assessment can lead to the introduction of irrelevant or redundant data, which might increase communication overhead as well as degrade global model performance. To address this problem, many research works have proposed participant selection strategies in FL \cite{liu2022gtg,tastan2024redefining,fan2022fair,wang2019measure,cho2022towards,goetz2019active,shi2023efficient,rokvicprivacy,shen2024research}. One prominent approach involves using Shapley values to evaluate each participant's contribution to the global model. Shapley values provide a theoretically sound and fair estimation of contributions; however, their computation is often prohibitively expensive due to the exponential complexity involved. Approximation methods, such as GTG-Shapley \cite{liu2022gtg}, which reconstructs models from gradient updates to efficiently estimate Shapley values, and ShapFed \cite{tastan2024redefining}, which combines Shapley values with class-specific metrics, have been proposed. Despite their efficiency, approximations may not always accurately capture contributions. VerFedSV \cite{fan2022fair} and the method proposed in \cite {wang2019measure} extend Shapley-based strategies to the VFL, tailoring them for vertically partitioned data to assess participant relevance. Another class of strategies evaluates model updates or training dynamics to guide participant selection. For example, the "Power-of-Choice" method \cite{cho2022towards} prioritizes participants with higher local losses to speed up convergence. Goetz et al. \cite{goetz2019active} introduced an active learning-based approach that selects clients based on the informativeness of their local updates. Similarly, Shi et al. \cite{shi2023efficient} modeled client selection as a multi-armed bandit problem, dynamically balancing exploration and exploitation. However, these methods operate during the training phase, meaning all participants are initially involved. This can result in higher communication and computational costs due to redundant or irrelevant parties being included. Considering this issue, some studies have focused on participant selection before the training phase by assessing local data quality and relevance. \cite{rokvicprivacy} introduces a privacy-preserving "lazy influence" technique, allowing participants to score data locally and share differentially private scores with the federation center. Another approach in \cite{shen2024research} proposes a two-stage data quality governance framework, combining local data assessment and outlier handling with an enhanced aggregation method (DQ-FedAvg) to improve both local and global model performance. However, these methods are only applicable to HFL. VFL requires different strategies due to its unique feature partitioning and the need to preserve feature-level privacy. 
\\\\
To the best of our knowledge, only two studies \cite{jiang2022vf,huang2023adaptive} have addressed participant selection in VFL prior to training. \cite{jiang2022vf} introduces \text\bf{VF-MINE}, which estimates mutual information (MI) between the features and labels of selected participants. Their method optimizes MI estimation using Fagin’s algorithm and proposes a group testing-based framework for participant selection. While \textit{VF-MINE} performs well, it assumes that all parties provide distinct features. This assumption does not hold in many real-world scenarios, where overlapping or redundant features are common across participants. \cite{huang2023adaptive} proposed \textit{VFLMG } which uses a gain-based greedy algorithm to select participants dynamically by balancing joint mutual information with the target and communication cost. However, dynamically determining the number of participants through such greedy algorithms may not always be ideal. It assumes that the immediate gain of adding a participant is the best criterion for decision-making, which can sometimes lead to short-sighted selection. To address the limitations of existing participant selection methods in VFL, we propose a novel approach that emphasizes participant relevance to the global model and computational efficiency. Our main contributions in this paper are:
\begin{itemize}
    \item We adopt a simpler computation of Spearman correlation using secure multi-party computation (SMPC) to measure the relevance of data provided by each participant unlike methods relying on computationally expensive mutual information estimation. Based on the feature correlation of the parties, each of them gets a score denoting their relevance.
    \item  We explicitly address the issue of redundancy among features across parties. By first identifying features with highly similar correlation patterns and then confirming redundancy through encrypted correlation checks, our method ensures that only parties with no or least redundant data are considered. Additionally, we propose a forward selection algorithm that iteratively selects participants based on their scores, which are recalculated after each selection round to account for redundancy. 
    \item We evaluated our approach on both regression and classification tasks using multiple datasets. Through comprehensive experiments, we compare our method against several baselines, demonstrating that our approach outperforms existing methods, even in scenarios with redundant or overlapping data. Notably, we show that using only 50\% of the participants achieves performance comparable to utilizing all participants.
\end{itemize}

\section{Prelimineries}

\begin{figure}[htbp]
    \centering
    \includegraphics[width=\columnwidth]{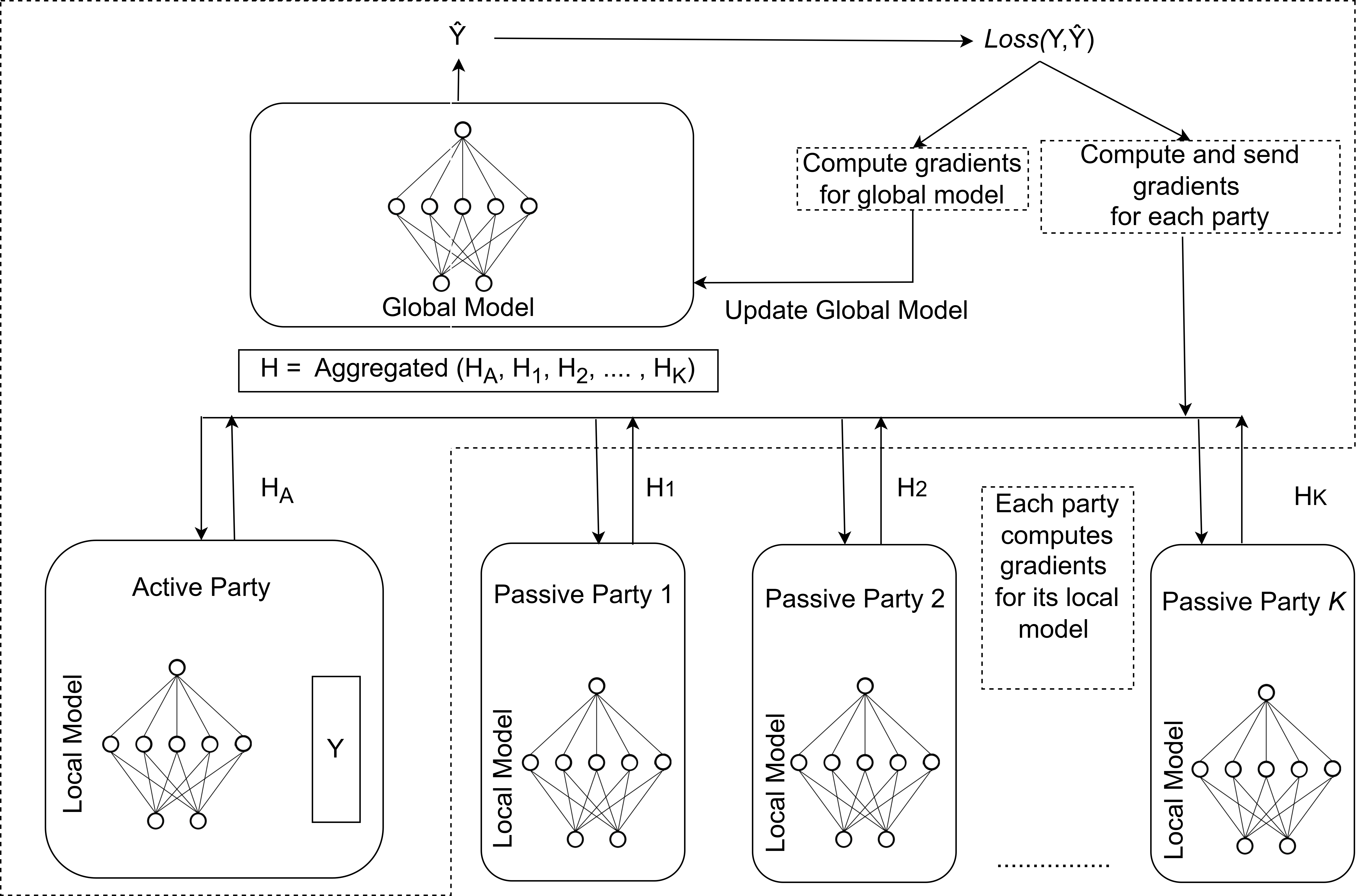} 
    \caption{Vertical Federated Learning Setup}
    \label{fig:vfl}
\end{figure}

Vertical Federated Learning is a collaborative machine learning paradigm where data is vertically partitioned across multiple parties based on features. Each party contributes a subset of features for the same set of data instances. In this setup, a single instance $\mathbf{x}_i \in \mathbb{R}^d$ is divided among $K$ parties, where $\mathbf{x}_{i,k} \in \mathbb{R}^{d_k}$ represents the subset of features held by party $k$, with $d_k$ as the dimensionality of its feature space. Among these, one party is designated as the \textit{active party}, which owns the labels $y_i$. The remaining parties, called \textit{passive parties}, provide feature contributions without access to the target labels (Figure \ref{fig:vfl}).

The data of the active party is represented as $\mathcal{D}_\text{active} = \{ (\mathbf{x}_{i,\text{active}}, y_i) \}_{i=1}^N$, where $N$ is the total number of instances. Each passive party $k$ holds a local data $\mathcal{D}_k = \{\mathbf{x}_{i,k}\}_{i=1}^N$. The goal of VFL is to train a joint model by securely utilizing the distributed features across parties while preserving data privacy. Each party processes its local features using a local model $f_k(\mathbf{x}_{i,k}; \boldsymbol{\theta}_k)$, parameterized by $\boldsymbol{\theta}_k$. The active party aggregates these outputs using a global model $h(\mathbf{z}; \boldsymbol{\phi})$, parameterized by $\boldsymbol{\phi}$, where $\mathbf{z}$ represents the combined outputs from all local models.

The overall objective of VFL is to minimize a global loss function over the aggregated outputs of all parties. The objective function can be expressed as:

\begin{multline}
\mathcal{L}_{\{1, \dots, K\}} = \frac{1}{N} \sum_{i=1}^N \ell\Big(h\big(f_1(\mathbf{x}_{i,1}; \boldsymbol{\theta}_1), \dots, \\
f_K(\mathbf{x}_{i,K}; \boldsymbol{\theta}_K); \boldsymbol{\phi}\big), y_i\Big),
\label{los1}
\end{multline}

where $\ell$ is the task-specific loss function, such as cross-entropy for classification or mean squared error for regression. During \textit{forward propagation}, each party computes $f_k(\mathbf{x}_{i,k})$ and sends the results to the active party, which aggregates them to compute the global loss. In \textit{backward propagation}, the active party calculates gradients of the global loss with respect to each $f_k(\mathbf{x}_{i,k})$ and sends these gradients back to the passive parties. Each party then updates its parameters $\boldsymbol{\theta}_k$, while the active party updates $\boldsymbol{\phi}$. 
\subsection{Motivating Example}
Vertical Federated Learning combines data from multiple parties to improve model performance using diverse feature sets. However, including all parties indiscriminately can lead to higher computational costs, slower training, and limited performance gains, especially in large-scale setups with many data parties.

\begin{figure}[htbp]
    \centering
    \includegraphics[width=\columnwidth]{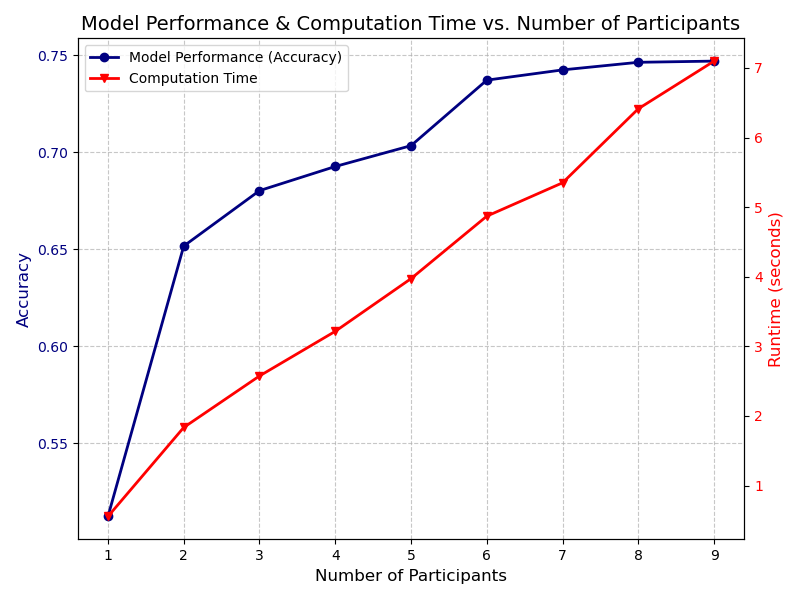} 
    \caption{Test accuracy and runtime with different number of passive parties}
    \label{fig:graph}
\end{figure}

To demonstrate this, we perform a small-scale experiment on the Wine Quality dataset from the UCI repository \cite{cortez2009modeling}. To simulate the VFL setting, the features of the dataset are split among 8 passive parties and 1 active party. It is observed from Figure \ref{fig:graph} that, adding more parties initially improved the model quality (F1-score), but the benefits leveled off after $5$ passive parties were selected. There was no significant improvement in the quality of the model once the F-Score $77.83\%$ was reached. Meanwhile, computational time increased steadily as more parties joined, showing the added overhead. Furthermore, practical scenarios often involve parties with redundant or irrelevant data. For example, some parties may hold features that overlap significantly with the active party or other passive parties, contributing little new information. In some cases, parties may have entirely irrelevant data that does not contribute to the learning task and can even degrade the model's performance by introducing noise or bias.  For instance, in banking, multiple financial institutions collaborating on credit risk assessment might each have access to a customer's income and transaction history, resulting in overlapping features, while some institutions might also contribute unrelated marketing data that does not improve the prediction of loan defaults. Including data from such parties not only wastes computational resources but may also harm the overall effectiveness of the model. This highlights the need for an effective participant selection mechanism in VFL. Choosing only the most relevant parties can keep or even improve model performance while cutting down on unnecessary computational costs.

\subsection{Participant Selection Problem in VFL}

The problem of participant selection in VFL is analogous to the feature selection problem in traditional machine learning, where the objective is to select only the most relevant features to improve model performance and efficiency. In the context of VFL, participants (or parties) holding important features are identified and selected to train a global model. Traditional feature selection methods like assessing correlations \cite{asuero2006correlation}, LASSO \cite{zhang2019feature}, and mutual estimator (MI) \cite{kraskov2004estimating} cannot be directly applied to VFL settings because feature information is distributed across multiple parties without being shared due to privacy constraints. As a result, selecting relevant participants requires designing something different approaches that achieve the goal without directly accessing raw feature values.

The goal is to select the top \( M \) most relevant participants from a total of \( K \) available parties, such that the global model trained using the selected subset achieves performance comparable to or better than the model trained using all \( K \) parties. Selecting only the most informative parties reduces computational costs and avoids redundancy without compromising the quality of the learned model.
The participant selection objective is to minimize this loss function by using data from a carefully chosen subset of parties, denoted as \( \mathcal{S} \subseteq \{1, 2, \dots, K\} \), where \( |\mathcal{S}| = M \). The corresponding loss function for the selected subset \( \mathcal{S} \) can be written as:
\[
\mathcal{L}_{\mathcal{S}} = \frac{1}{N} \sum_{i=1}^N \ell\Big(h\big(\{f_k(\mathbf{x}_{i,k}; \boldsymbol{\theta}_k) \mid k \in \mathcal{S}\}; \boldsymbol{\phi}\big), y_i\Big).
\]
The objective can then be formulated as:
\[
\text{Find } \mathcal{S} \text{ such that } |\mathcal{S}| = M, \text{ and } \mathcal{L}_{\{1, \dots, K\}} - \mathcal{L}_{\mathcal{S}} < \varepsilon.
\]

This formulation leverages the observation that not all parties contribute equally to the learning task. Some parties may have features that are highly redundant with others, while others may hold irrelevant or noisy features that degrade the model's performance. By ranking the parties based on their contributions to the global model, it becomes possible to identify and retain only the most relevant ones.

In summary, the participant selection problem in VFL seeks to identify a subset of \( M \) participants that collectively maximize the performance of the global model while minimizing unnecessary computational overhead. This problem is challenging due to the lack of direct access to individual feature information, privacy constraints, and the potential redundancy or irrelevance of certain features held by different parties.

\section{Proposed Method}

\begin{figure}[htbp]
    \centering
    \includegraphics[width=\columnwidth]{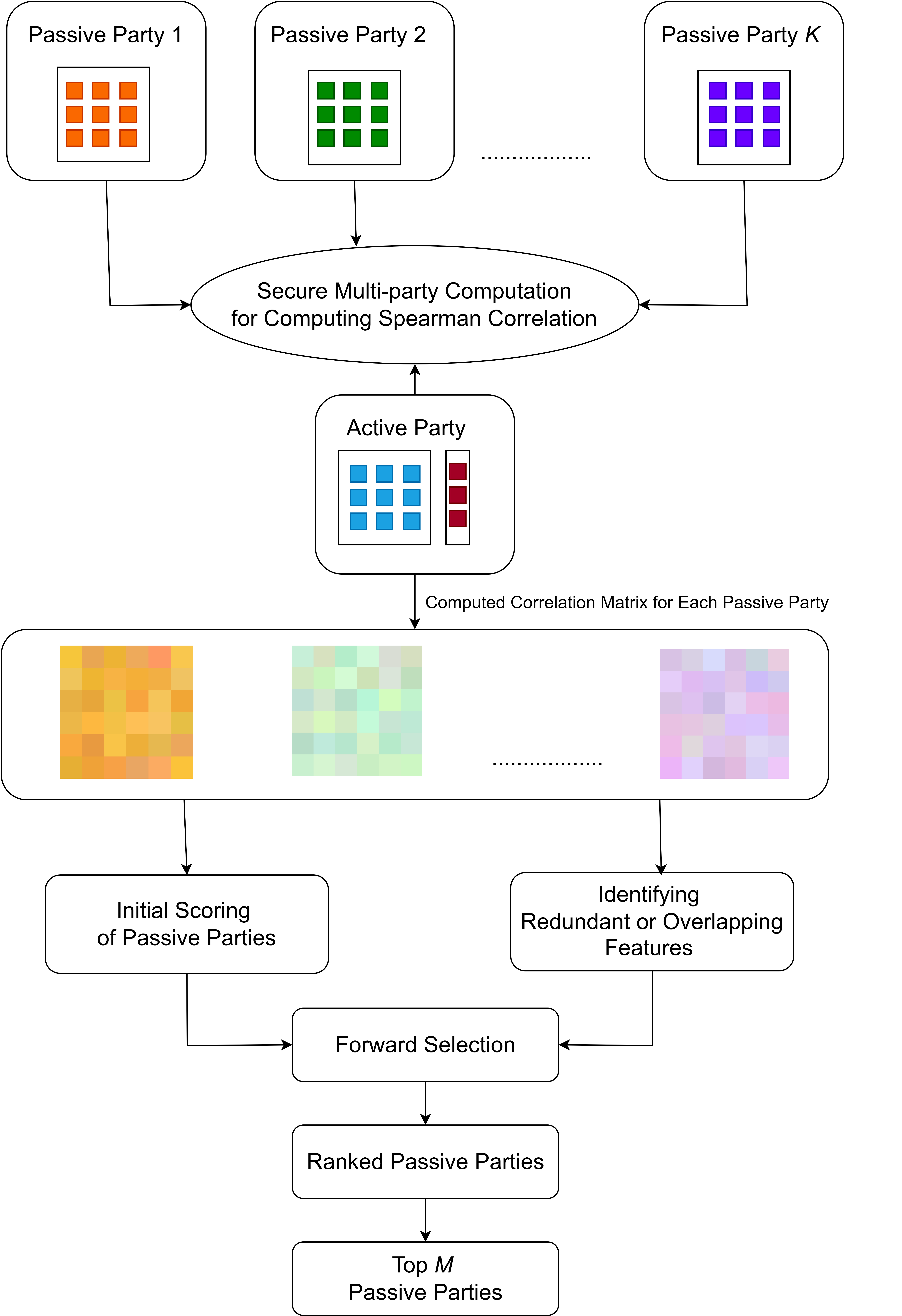} 
    \caption{VFL-RPS Workflow}
    \label{fig:method}
\end{figure}

In this section, we present our proposed method for selecting relevant participants in VFL. In a VFL setting, the passive parties possessing features that have high relevance to the learning task are ideally the important ones. In a VFL setting, passive parties with features having high relevance to the learning task are preferred, particularly those with low correlation to the features of active party  but high correlation with the target variable, as they provide complementary information and give more insights in general. Additionally, passive parties with the least overlapping or redundant data must be selected, as redundant features do not contribute new information and can increase computational costs while potentially degrading model performance. Keeping this in mind, our proposed method selects relevant participants following two key steps (Figure \ref{fig:method}. In the first step, the active party securely calculates the correlation between its own features and those of each passive party, as well as the correlation between each passive party’s features and the target variable, which is only possessed by the active party. Based on the obtained correlations, we identify redundant or overlapping features between the active party and each passive party, as well as among passive parties by checking if the correlations exceed a specific defined threshold (eg: 0.9). This gives information about what the unique and redundant features among the passive parties. In next step, we assign a relevance score to each passive party based on the correlations computed in the earlier step. A forward selection strategy is then applied, iteratively selecting the highest-scoring passive party while recalculating scores for the remaining parties accounting for the redundant features among them.
\subsection{Secure Correlation Computation and Identifying Redundancy}

To assess the relevance of features in passive parties, we securely compute the Spearman correlation coefficient using secure multi-party computation protocol(SMPC) \cite{cramer2015secure}. Let the active party hold the dataset \(\mathcal{D}_a = \{\mathbf{x}_{i,a}, y_i\}_{i=1}^N\), where \(\mathbf{x}_{i,a} \in \mathbb{R}^d\) are features and \(y_i\) is the target variable for the \(i\)-th sample. Each passive party \(P_k\) holds the dataset \(\mathcal{D}_k = \{\mathbf{x}_{i,k}\}_{i=1}^N\), where \(\mathbf{x}_{i,k} \in \mathbb{R}^{d_k}\). The goal is to compute the correlation between the active party’s features (including the target) and those of each passive party.

The SPCCAdv algorithm, proposed by \cite{hong2018secure}, uses a random matrix-based method to securely compute Pearson correlation coefficients between two variables. In SPCCAdv, each party standardizes its data by subtracting the mean and dividing by the standard deviation, ensuring that the normalized data has a mean of 0 and a standard deviation of 1. This prevents the disclosure of sensitive statistics such as mean and standard deviation, which could otherwise allow inference about the raw data. Using this standardized data, SPCCAdv computes the Pearson correlation through a secure scalar product, as follows:
\[
\text{Pearson}(\bar{X}, \bar{Y}) = \frac{\sum_{i=1}^k \bar{x}_i \cdot \bar{y}_i}{k}
\]
where \(\bar{X}\) and \(\bar{Y}\) are the normalized datasets.

While SPCCAdv ensures privacy during the computation of Pearson correlation, its reliance on linear relationships makes it unsuitable for monotonic relationships. To address this, we adapt SPCCAdv for Spearman correlation, which is computed as the Pearson correlation on rank-transformed data. By replacing raw data with their ranks, we preserve monotonic relationships, making the method applicable in a broader range of scenarios.

Our adaptation involves three steps:
(1) Each party independently transforms its data into ranks and normalizes the ranks to have a mean of 0 and a standard deviation of 1.
(2) The normalized ranks are securely exchanged using the random matrix-based secure scalar product method.
(3) The secure scalar product of the normalized ranks is computed to obtain the Spearman correlation:
\begin{equation}
   \text{Spearman}(\bar{R}_X, \bar{R}_Y) = \frac{\sum_{i=1}^k \bar{r}_{X,i} \cdot \bar{r}_{Y,i}}{k}
\end{equation}

This approach inherits the privacy-preserving guarantees of SPCCAdv while extending its applicability. The detailed pseudocode is presented in Algorithm~\ref{alg:smpc_spearman}.

\begin{algorithm}[h!]
\caption{SPCC\_Spearman: Secure Spearman Correlation Computation}
\label{alg:smpc_spearman}
\begin{algorithmic}[1]
\Require \( X \) and \( Y \) are \( k \)-length data owned by Alice and Bob, respectively.
\Ensure Spearman correlation \(\rho\) between \(X\) and \(Y\).
\State \textbf{Alice and Bob:} Transform \(X\) and \(Y\) into ranks \(R_X\) and \(R_Y\). Standardize ranks to \(\bar{R}_X\) and \(\bar{R}_Y\).
\State Assume that Alice and Bob share a \( k \times \frac{k}{2} \) random matrix \( A = [a_{i,j}] \).
\Statex
\State \textbf{Alice: Encrypt \(\bar{R}_X\):}
\begin{enumerate}
    \item Generate a random vector \( R \) of length \(\frac{k}{2}\).
    \item Compute \( Z = \bar{R}_X + AR \).
    \item Send \( Z \) to Bob.
\end{enumerate}
\Statex
\State \textbf{Bob: Encrypt \(\bar{R}_Y\):}
\begin{enumerate}
    \item Compute \( s = Z^\top \bar{R}_Y \).
    \item Compute \( V = A^\top \bar{R}_Y \).
    \item Send \( s \) and \( V \) to Alice.
\end{enumerate}
\Statex
\State \textbf{Alice: Compute Spearman Correlation:}
\begin{enumerate}
    \item Compute \( s' = V^\top R \).
    \item Compute the scalar product: \(\bar{R}_X \cdot \bar{R}_Y = s - s'\).
    \item Compute the Spearman correlation:
    \[
    \rho = \frac{\bar{R}_X \cdot \bar{R}_Y}{k}
    \]
\end{enumerate}
\end{algorithmic}
\end{algorithm}
In this protocol, Alice represents the active party, responsible for holding the target variable and a subset of features, while Bob represents a passive party contributing additional feature sets. The protocol is specifically tailored for VFL settings involving a single active party (Alice) securely computing correlations with multiple passive parties (Bob$_1$, Bob$_2$, ..., Bob$_n$). Once the correlations are computed, stores the correlation matrix \( \mathbf{C}_k \) for each passive party structured as follows:

\begin{small}
\begin{equation}
\mathbf{C}_k=
\begin{bmatrix}
\rho(\mathbf{x}_{a,1}, \mathbf{x}_{k,1}) & \dots & \rho(\mathbf{x}_{a,1}, \mathbf{x}_{k,d_k}) \\
\rho(\mathbf{x}_{a,2}, \mathbf{x}_{k,1}) & \dots & \rho(\mathbf{x}_{a,2}, \mathbf{x}_{k,d_k}) \\
\vdots & \vdots & \ddots & \vdots \\
\rho(\mathbf{x}_{a,d}, \mathbf{x}_{k,1}) &  \dots & \rho(\mathbf{x}_{a,d}, \mathbf{x}_{k,d_k}) \\
\hline
\rho(y, \mathbf{x}_{k,1}) &  \dots & \rho(y, \mathbf{x}_{k,d_k})
\end{bmatrix}
\end{equation}
\end{small}
\noindent where, each \( \rho(\mathbf{x}_{a,i}, \mathbf{x}_{k,j}) \) represents the Spearman correlation between the \( i \)-th active party feature and the \( j \)-th feature of passive party \( k \). In the last row, \( \rho(y, \mathbf{x}_{k,j}) \), represents the Spearman correlation of \( j \)-th feature of the passive party \( k \) with the target variable. This structure allows the active party to analyze the relationships between features across all parties.

After computing the correlation matrices for all passive parties, the active party identifies overlapping and redundant features through a two-step process. First, it scans each correlation matrix, examining each column to determine the highest absolute correlation value. If a feature exhibits a correlation above a predefined threshold (e.g., 0.9) with any active party feature, it is flagged as overlapping with the active party.

Once redundancy with the active party is identified, redundancy among passive parties is detected. The active party compares the correlation matrices \( \mathbf{C}_i \) and \( \mathbf{C}_j \) of two passive parties \( P_i \) and \( P_j \). A feature \( x_{i,m} \) from \( P_i \) and \( x_{j,n} \) from \( P_j \) are considered redundant if their correlation patterns with all active party features and the target are highly similar. Specifically, redundancy is flagged when:

\begin{equation}
\| \mathbf{C}_i(:,m) - \mathbf{C}_j(:,n) \|< \delta
\end{equation}

where \( \mathbf{C}_i(:,m) \) and \( \mathbf{C}_j(:,n) \) are the correlation vectors of features \( x_{i,m} \) and \( x_{j,n} \), containing their correlations with all active party features and the target, and \( \delta \) is a small threshold.

To confirm redundancy, the active party computes the direct correlation between \( x_{i,m} \) and \( x_{j,n} \). If:

\begin{equation}
|\rho(x_{i,m}, x_{j,n})| > \tau
\end{equation}

where \( \tau \) is a predefined high threshold (e.g., 0.95), the features are marked as redundant.

\subsection{Participant Scoring and Forward Selection}
After identifying the redundancies, each passive party \( P_k \) is assigned a relevance score based on the correlation of its features with both the active party’s features and the target variable. This scoring mechanism ensures that features contributing uniquely to the target are prioritized while redundant features are down-weighted. For each feature \( f \) in the set of unique features (not overlapping with active party features) \( \mathcal{F}_k \) of a passive party \( P_k \), the relevance score is computed as:
\[
\text{Score}(P_k) = \sum_{f \in \mathcal{F}_k} \sum_{i=1}^{d} (1 - |\rho(f, x_{a,i})|) \cdot |\rho(f, y)|
\]
where:
\begin{itemize}
    \item \( \rho(f, x_{a,i}) \) is the Spearman correlation between the feature \( f \) and the \( i \)-th feature of the active party,
    \item \( \rho(f, y) \) is the Spearman correlation between the feature \( f \) and the target variable \( y \),
    \item \( d \) is the total number of features in the active party.
\end{itemize}

The total score for a passive party \( P_k \) is then computed as the sum of the scores for all its features:
\begin{equation}
\text{Score}(P_k) = \sum_{f \in \mathcal{F}_k} \text{Score}(f).
\label{score}
\end{equation}

Once all the passive parties are initially scored based on their relevance, we then apply a forward selection algorithm that identifies the top \( M \) passive parties by iteratively selecting the most relevant parties and recalculating the scores of the remaining ones to account for redundancy among the passive parties. At each iteration, the passive party with the highest score is added to the selected set \( \mathcal{S} \). Once a party \( P_k \) is selected, the scores of the remaining parties are recalculated to account for redundancy. If a feature \( f \in \mathcal{F}_j \) in a party \( P_j \) is found to be redundant with a feature in \( P_k \) (identified in the previous step), its contribution to the score of \( P_j \) is set to zero. The score for \( P_j \) is recalculated using the Equation \ref{score} like before. This iterative process continues until either the desired number of participants \( M \) is selected or can be used to rank all the passive parties.
\begin{algorithm}[h!]
\caption{Forward Selection Algorithm}
\label{alg:forward_selection}
\begin{algorithmic}[1]
\Require Set of passive parties \(\{P_1, P_2, \dots, P_K\}\), number of participants \( M \)
\Ensure Selected participant set \( \mathcal{S} \)
\State Initialize \( \mathcal{S} \gets \emptyset \)
\State Compute initial \(\text{Score}(P_k)\) for all parties \( P_k \) using:
\[
\text{Score}(P_k) = \sum_{f \in \mathcal{F}_k} \sum_{j=1}^{d} (1 - |\rho(f, x_{a,j})|) \cdot |\rho(f, y)|
\]
\While{\(|\mathcal{S}| < M\)}
    \State Select the party with the highest score:
    \[
    P_k \gets \arg\max_{P \notin \mathcal{S}} \text{Score}(P)
    \]
    \State Add \( P_k \) to \( \mathcal{S} \)
    \ForAll{\( P_j \notin \mathcal{S} \)}
        \ForAll{\( f \in \mathcal{F}_j \)}
            \If{\( f \) redundant with any feature in \( P_k \)}
                \State \( \text{Score}(f) = 0 \)
            \EndIf
        \EndFor
        \State Recalculate \(\text{Score}(P_j)\)
    \EndFor
\EndWhile
\State \Return \( \mathcal{S} \)
\end{algorithmic}
\end{algorithm}

\section{Experimental Setup}
The experiments are designed to answer two key questions: (1) Can selecting a smaller subset of relevant participants achieve performance comparable to or better than involving all parties? (2) How does our approach compare to existing baselines in terms of predictive performance and computational efficiency?

\subsection{Datasets and VFL Training}
We evaluate our proposed method in six publicly available datasets - three for regression tasks and three for classification tasks. 
\begin{table}[h]
\centering
\caption{Summary of datasets used for evaluation}
\label{tab:datasets}
\begin{tabular}{lcc}
\hline
\textbf{Dataset} & \textbf{Samples} & \textbf{Features} \\
\hline
\( D_1 \) California Housing \cite{pace1997sparse}  & 16005  & 12  \\
\( D_2 \) Steel Fatigue Strength \cite{agrawal2014exploration} & 437  & 26  \\
\( D_3 \) Wine Quality (Regression) \cite{cortez2009modeling} & 5329  & 12  \\
\( D_4 \) Credit Card Default \cite{yeh2009comparisons} & 37354  & 23  \\
\( D_5 \) Breast Cancer \cite{street1993nuclear} & 569  & 30  \\
\( D_6 \) Wine Quality (Classification) \cite{cortez2009modeling} & 4898  & 11  \\
\hline
\end{tabular}
\end{table}

Each dataset is split into training and testing sets using an 80/20 ratio. To simulate the VFL scenario, one party is designated as the \textbf{active party}, holding the target labels and a subset of features, while the remaining features are distributed among $K$ \textbf{passive parties}. The specific distribution of features varies based on the experimental configuration, as described below.

\subsection{Configurations}

To evaluate the robustness of the proposed approach, we consider three configurations that represent varying levels of complexity and challenges in VFL:

\begin{itemize}
    \item \textbf{Basic Configuration (Non-Overlapping Features):} Each passive party holds a unique, non-overlapping subset of features. This represents the simplest case, where all parties contribute distinct information.
    
    \item \textbf{Overlapping Features Configuration:} Passive parties may share overlapping features, either with one another or with the active party. This scenario reflects realistic settings where redundancy across organizations or databases can occur.
    
    \item \textbf{Irrelevant (Noisy) Features Configuration:} Some passive parties are given randomly generated, irrelevant features. This setup mimics scenarios where certain parties contribute noisy or unhelpful data, potentially degrading model performance if not excluded.
\end{itemize}

\subsection{Baselines}
We evaluate our participant selection method \textbf{VFL-RPS} with some baseline methods such as \textbf{ALL: } selects all \( K \) passive parties into the training process in VFL, providing an upper-bound benchmark for model performance. \textbf{ACTIVE\_ONLY: } utilizes only the data from the active party, offering a lower-bound reference for comparison. \textbf{RANDOM: } randomly chooses \( M \) passive parties without evaluating their relevance to the task from all \( M \) parties. \textbf{LASSO: } employs \(\ell_1\)-regularized linear regression, a feature selection technique from scikit-learn, to rank features centrally. High-importance participants are then selected based on the absolute sum of the coefficients assigned to their features. \textbf{VFLMG: } computes a participant's gain based on the joint mutual information of their features with the target labels and uses a greedy algorithm to maximize these gain values. We selected \textit{VFLMG} as a baseline from the existing methods since it is the most recent method proposed on the participant selection problem in VFL and outperforms the other method \textit{VF-MINE}.

\subsection{Models and Training}

We use vertically federated linear regression for regression tasks and vertically federated logistic regression for classification tasks. The models are trained using standard VFL protocol, where parties compute intermediate outputs locally and share encrypted updates with the active party \cite{liu2024vertical}. Training is configured with a fixed learning rate (0.01) until convergence (epoch = 1000). The VFL model is trained using the local data of top $M$ selected participants from all baselines and our method. The computational cost of the selection methods is included in our runtime analysis.

\subsection{Evaluation Metrics}

The proposed method is evaluated by training the VFL model with the selected participants and using standard metrics for regression (RMSE, $R^2$) and classification (Accuracy, F1 Score) to measure model performance, along with selection time to assess computational efficiency. As a comparison, Gini Importance—a metric \cite{nembrini2018revival} for feature relevance is also used. It is computed by training a centralized decision tree model on the combined data and aggregating feature importances for each passive party. This is done to evaluate how well the participant rankings produced by the proposed selection method align with those derived from Gini Importance, providing a reference point.

\section{Results}
Table \ref{table:overview} shows an overview of the performance of our proposed method for participant selection in VFL compared to other baseline methods.
\begin{table*}[t]
\centering
\begin{tabular}{|l|ccc|ccc|}
\hline
\multicolumn{1}{|c|}{}                            & \multicolumn{3}{c|}{\textbf{\begin{tabular}[c]{@{}c@{}}Regression\\ MSE(↓)\end{tabular}}}                                                                              & \multicolumn{3}{c|}{\textbf{\begin{tabular}[c]{@{}c@{}}Classification\\ Accuracy(↑)\end{tabular}}}                                                                     \\
\multicolumn{1}{|c|}{\multirow{-2}{*}{\textbf{Baselines}}} & \begin{tabular}[c]{@{}c@{}}D1\\ K=5, M=3\end{tabular} & \begin{tabular}[c]{@{}c@{}}D2\\ K=8, M=4\end{tabular} & \begin{tabular}[c]{@{}c@{}}D3\\ K=4, M=2\end{tabular} & \begin{tabular}[c]{@{}c@{}}D4\\ K=6, M=3\end{tabular} & \begin{tabular}[c]{@{}c@{}}D5\\ K=8, M=4\end{tabular} & \begin{tabular}[c]{@{}c@{}}D6\\ K=4, M=2\end{tabular} \\ \hline
ALL                                     & 0.32                                                  & 0.03                                                  & 0.56                                                  & 0.72                                                  & 0.96                                                  & 0.73                                                  \\
ACTIVE\_ONLY                             & 0.51                                                  & 0.90                                                  & 0.76                                                  & 0.61                                                  & 0.91                                                  & 0.67                                                  \\ \hline
RANDOM                                 & 0.35                                                  & 0.15                                                  & 0.62                                                  & 0.69                                                  & 0.95                                                  & 0.70                                                  \\
LASSO                                    & \textbf{0.33}                           & \textbf{0.08}                           & 0.71                                                  & 0.71                                                  & 0.91                                                  & 0.68                                                  \\
VFLMG                                    & 0.34                                                  & 0.09                                                  & 0.70                                                  & 0.68                                                  & \textbf{0.98}                           & 0.68                                                  \\
\textbf{VFL-RPS}                                   & \textbf{0.33}                           & \textbf{0.08}                          & \textbf{0.57}                           & \textbf{0.72}                           & \textbf{0.98}                           & \textbf{0.73}                           \\ \hline
\end{tabular}
\caption{Performance comparison of different baseline methods on regression and classification tasks. The table reports MSE and Accuracy for classification across six datasets (D1–D6), each configured with different values of K (number of total passive parties) and M (number of selected passive parties). Our proposed method, \textbf{VFL-RPS}, achieves the best or comparable performance across most datasets, as highlighted in red. Notably, in most cases \textbf{VFL-RPS} outperforms other baseline methods such as RANDOM, LASSO, VFLMG}
\label{table:overview}
\end{table*}
 
\begin{table*}[h]
\centering
\scriptsize
\setlength{\tabcolsep}{4pt} 
\renewcommand{\arraystretch}{0.95}
\begin{tabular}{|l|c|c|c|p{6cm}|} 
\hline
\textbf{Baselines}         & \textbf{MSE (↓)} & \textbf{R² (↑)} & \textbf{Selection Time (s)} & \textbf{Rankings (Top M=3)}                     \\ \hline
ALL           & 0.32             & 0.67            & --                          & --                                             \\ 
ACTIVE ONLY    & 0.51             & 0.48            & --                          & --                                             \\ \hline
RANDOM         & 0.35             & 0.65            & --                          & --                                             \\ 
LASSO          & \textbf{0.33} & \textbf{0.67} & 0.1042                      & \textbf{host1}, \textbf{host2}, \textbf{host3}, host4, host5 \\ 
VFLMG          & 0.34             & 0.66            & 3.6350                      & \textbf{host1}, \textbf{host3}, \textbf{host4}, host2, host5 \\ 
\textbf{VFL-RPS}         & \textbf{0.33} & \textbf{0.67} & 1.8027                      & \textbf{host5}, \textbf{host2}, \textbf{host1}, host4, host3 \\ 
Gini Importance Score & --              & --              & --                          & \textbf{host5}, \textbf{host1}, \textbf{host2}, host4, host3 \\ \hline
\end{tabular}
\caption{Performance comparison on California Housing Dataset (Basic Configuration). The active party holds a subset of the features along with the target, while the remaining features are uniquely distributed among 5 passive parties (hosts1-5). This configuration ensures no redundancy feature distribution, representing an idealized setup. VFL-RPS selects the top 3 parties (\{5, 2, 1\}), corresponding to 50\% of the total parties, achieving an MSE of 0.33 and R² of 0.67, matching LASSO and outperforming VFLMG. Notably, VFL-RPS and Gini Importance rankings agree on the top 3 hosts, validating the relevance of the selected parties.}
\label{tab:california_basic}
\end{table*}

\begin{table*}[t]
\centering
\scriptsize
\setlength{\tabcolsep}{4pt} 
\renewcommand{\arraystretch}{0.95}
\begin{tabular}{|l|c|c|c|p{6cm}|} 
\hline
\textbf{Baselines}         & \textbf{MSE (↓)} & \textbf{R² (↑)} & \textbf{Selection Time (s)} & \textbf{Rankings (Top M=3)}                     \\ \hline
ALL           & 0.32             & 0.67            & --                      & --                                             \\ 
ACTIVE ONLY    & 0.51             & 0.48            & --                          & --                                             \\ \hline
RANDOM         & 0.38             & 0.62            & --                      & --                                             \\ \
LASSO          & 0.41             & 0.59            & 0.1109                      & \textbf{host6}, \textbf{host1}, \textbf{host2}, host3, host4, host5, host7, host8 \\ 
VFLMG          & 0.34             & 0.66            & 5.3796                      & \textbf{host7}, \textbf{host1}, \textbf{host3}, host4, host2, host6, host5, host8 \\ 
\textbf{VFL-RPS}         & \textbf{0.33} & \textbf{0.67} & 3.8024                      & \textbf{host5}, \textbf{host2}, \textbf{host1}, host4, host7, host3, host6, host8 \\ 
Gini Importance Score & --              & --              & --                          & \textbf{host5}, \textbf{host1}, \textbf{host2}, host6, host8, host4, host7, host3 \\ \hline
\end{tabular}
\caption{Performance comparison and rankings for California Housing (Overlapping Features Configuration). To simulate redundancy, the total number of passive parties is increased to 8 (host6-8) by duplicating features across parties. VFL-RPS continues to select 3 parties, representing 50\% of the original ideal configuration. It achieves an MSE of 0.33 and R² of 0.67, outperforming LASSO and VFLMG. Notably, VFL-RPS ranks and selects \{host5, host2, host1\}, which aligns with Gini Importance rankings for the top 3 hosts, demonstrating its ability to exclude redundant parties effectively.}
\label{tab:california_overlapping}
\end{table*}
 
\begin{table*}[t]
\centering
\scriptsize
\setlength{\tabcolsep}{4pt} 
\renewcommand{\arraystretch}{0.95}
\begin{tabular}{|l|c|c|c|p{6cm}|} 
\hline
\textbf{Baselines}         & \textbf{MSE (↓)} & \textbf{R² (↑)} & \textbf{Selection Time (s)} & \textbf{Rankings (Top M=3)}                     \\ \hline
ALL            & 0.32             & 0.67            & --                      & --                                             \\ \
ACTIVE ONLY    & 0.51             & 0.48            & --                          & --                                             \\ \hline
RANDOM         & 0.40             & 0.59            & --                      & --                                             \\ 
LASSO         & 0.41             & 0.59            & 0.1125                      & \textbf{host1}, \textbf{host2}, \textbf{host3}, host4, host5, host6, host7, host8 \\ 
VFLMG         & 0.42             & 0.57            & 6.1716                      & \textbf{host1}, \textbf{host4}, \textbf{host2}, host6, host5, host7, host3, host8 \\ 
\textbf{VFL-RPS}         & \textbf{0.33} & \textbf{0.67} & 3.7724                      & \textbf{host8}, \textbf{host3}, \textbf{host1}, host7, host5, host4, host2, host6 \\ 
Gini Importance Score & --              & --              & --                          & \textbf{host8}, \textbf{host1}, \textbf{host3}, host7, host2, host6, host4, host5 \\ \hline
\end{tabular}
\caption{Performance comparison and rankings on California Housing (Irrelevant features configuration). To simulate the presence of noise, some parties (host2, host4, host6) hold randomly generated irrelevant features. The total number of passive parties remains 8, and VFL-RPS selects 3 parties (\{host8, host3, host1\}), representing 50\% of the total parties. Notably, parties with irrelevant data (host2, host4, host6) are ranked lowest and excluded from selection, demonstrating VFL-RPS's ability to effectively identify and exclude noisy participants while achieving an MSE of 0.33 and R² of 0.67, outperforming LASSO and VFLMG.}
\label{tab:california_unrelated}
\end{table*}
\vspace{5mm}
\begin{table*}[!t]
\centering
\scriptsize
\setlength{\tabcolsep}{4pt} 
\renewcommand{\arraystretch}{0.95}
\begin{tabular}{|l|c|c|c|p{6cm}|} 
\hline
\textbf{Baselines}         & \textbf{F1 Score (↑)} & \textbf{Accuracy (↑)} & \textbf{Selection Time (s)} & \textbf{Rankings (Top M=3)}                     \\ \hline
ALL            & 0.73             & 0.72            & --                      & --                                             \\ 
ACTIVE ONLY    & 0.61             & 0.63            & --                      & --                                             \\ \hline
RANDOM         & 0.69             & 0.69            & --                  & --                                             \\ 
LASSO          & 0.71             & 0.71            & 0.6031                  & \textbf{host1}, \textbf{host2}, \textbf{host3}, host4, host5, host6 \\ 
VFLMG          & 0.71             & 0.71            & 325.2042                & \textbf{host2}, \textbf{host3}, \textbf{host1}, host4, host5, host6 \\ 
VFL-RPS  & \textbf{0.72} & \textbf{0.72} & 17.7166                 & \textbf{host1}, \textbf{host4}, \textbf{host6}, host2, host3, host5 \\ 
Gini Importance Score & --              & --              & --                      & \textbf{host1}, \textbf{host4}, \textbf{host3}, host6, host5, host2 \\ \hline
\end{tabular}
\caption{Performance comparison on Credit Card Default Dataset (Basic Configuration). In a similar manner, the dataset features are split into 1 active party and 6 passive parties. We can observe that, our method VFL-RPS outperforms RANDOM, LASSO \& VFLMG and its ranking is more aligned with the gini importance score based ranking compared to the other baselines.}
\label{tab:credit_basic}
\end{table*}

\begin{table*}[!t]
\centering
\scriptsize
\setlength{\tabcolsep}{4pt} 
\renewcommand{\arraystretch}{0.95}
\begin{tabular}{|l|c|c|c|p{6cm}|} 
\hline
\textbf{Baselines}         & \textbf{F1 Score (↑)} & \textbf{Accuracy (↑)} & \textbf{Selection Time (s)} & \textbf{Rankings (Top M=3)}                     \\ \hline
ALL            & 0.72      & 0.72     & --                      & --                                             \\ 
ACTIVE ONLY    & 0.61             & 0.63            & --                      & --                                             \\ \hline
RANDOM         & 0.70      & 0.70      & --                  & --                  \\ 
LASSO          & \textbf{0.72}       & \textbf{0.72}      & 5.5644                  & \textbf{host7}, \textbf{host8}, \textbf{host4}, host6, host5, host1, host2, host3, host9 \\  
VFLMG          & 0.70       & 0.68      & 383.1599                & \textbf{host2}, \textbf{host3}, \textbf{host4}, host1, host5, host6, host7, host8, host9 \\ 
VFL-RPS  & \textbf{0.72} & \textbf{0.72} & 19.2039                 & \textbf{host1}, \textbf{host9}, \text{host6}, host4, host3, host5, host2, host7, host8 \\  
Gini Importance Score & --              & --              & --                      & \textbf{host1}, \textbf{host7}, \textbf{host3}, host4, host8, host2, host6, host5, host9 \\ \hline
\end{tabular}
\caption{Performance comparison on Credit Card Default Dataset (Overlapping Features Configuration). The same setting in Table \ref{tab:credit_basic} is followed but in this case, to simulate redudancy, passive parties hosts7-9 are added which have redundant features from previous parties. After selection of top 3 passive parties, LASSO and VFLMG are observed to show comprabale performance.}
\label{tab:credit_overlapping}
\end{table*}

\begin{table*}[!t]
\centering
\scriptsize
\setlength{\tabcolsep}{4pt} 
\renewcommand{\arraystretch}{0.95}
\begin{tabular}{|l|c|c|c|p{6cm}|} 
\hline
\textbf{Baselines}         & \textbf{F1 Score (↑)} & \textbf{Accuracy (↑)} & \textbf{Selection Time (s)} & \textbf{Rankings (Top M=3)}                     \\ \hline
ALL            & 0.73      & 0.72     & --                      & --                                             \\ 
ACTIVE ONLY    & 0.61            & 0.63           & --                      & --                                             \\ \hline
RANDOM         & 0.69     & 0.69    & --                 & --                  \\ 
LASSO          & \textbf{0.73}      & \textbf{0.72}    & 1.0504                  & \textbf{host1}, \textbf{host2}, \textbf{host3}, host4, host5, host6, host7, host8, host9 \\ 
VFLMG          & \textbf{0.73}      & \textbf{0.72}     & 409.9671                & \textbf{host3}, \textbf{host4}, \textbf{host2}, host6, host7, host8, host1, host5, host9 \\ 
VFL-RPS  & \textbf{0.73}   & \textbf{0.72}  & 19.2945         & \textbf{host2}, \textbf{host6}, \textbf{host8}, host3, host4, host7, host1, host5, host9 \\ 
Gini Importance Score & --              & --              & --                      & \textbf{host2}, \textbf{host6}, \textbf{host4}, host8, host7, host3, host1, host5, host9 \\ \hline
\end{tabular}
\caption{Performance comparison on Credit Card Default Dataset (Irrelevant Features Configuration). In this case, three additional passive parties, hosts(1,5,9) are added to the setting of \ref{tab:credit_basic}, and the additional parties are assigned features randomly generated to simulate irrelevant features for the learning task. From the results, it is observed that even tho model performance wise, LASSO, VFLMG \& VFL-RPS give comparable results, only VFLMG \& VFL-RPS can identify the parties contributing irrelevant features but VFL-RPS does it in a shorter time. Our method ranks the those three parties, hosts(1,5,9) as the last 3 which also aligns with the gini importance-based ranking.}
\label{tab:credit_unrelated}
\end{table*}

To further evaluate the effectiveness of our proposed method, we assess its performance in scenarios where passive parties contain highly overlapping or redundant features (overlapping features configuration) and where some parties hold entirely irrelevant features (unrelated features configuration). Due to page constraints, we only present results for one dataset from each task—regression (Tables \ref{tab:california_basic}, \ref{tab:california_overlapping} \& \ref{tab:california_unrelated}) and classification (Tables \ref{tab:credit_basic}, \ref{tab:credit_overlapping} and \ref{tab:credit_unrelated}). However, we have observed similar trends across all datasets, confirming the consistency and robustness of our method in selecting the most informative parties across different configurations.
\section{Discussion and Future Work}
We proposed a novel participant selection method \textbf{VFL-RPS} for vertically federated settings where feature or data information from data parties are not shared due to privacy constraints. It also considers cases where parties might have redundant features among them as well as features that are irrelevant to the learning task. Experimental results demonstrate that our method effectively selects the most informative participants while also identifying parties with irrelevant data, ensuring a more efficient and high-performing global model. 
As our method is specifically designed for numerical tabular data, it may not generalize well to unstructured data such as images or text. 
To improve generalizability, we aim to incorporate a combination of correlation analysis techniques tailored to different data types. For example, using Spearman correlation for ranked numerical data, Chi-square tests for categorical variables, and distance correlation for capturing nonlinear dependencies. This multi-metric approach would allow our method to adapt to a broader range of data types, improving participant selection across diverse VFL applications. However,  our method still remains well-suited for many real-world cases where numerical tabular data with ranked features is exists.

\bibliographystyle{IEEEtran}
\bibliography{references.bib}




\end{document}